\documentclass{article}

\usepackage[final]{tackling_climate_workshop_style}

\usepackage[utf8]{inputenc} 
\usepackage[T1]{fontenc}    
\usepackage{hyperref}       
\usepackage{url}            
\usepackage{booktabs}       
\usepackage{amsfonts}       
\usepackage{nicefrac}       
\usepackage{microtype}      
\usepackage{graphicx}
\usepackage{float}

\title{AI-based Waste Mapping for Addressing Climate-Exacerbated Flood Risk}

\author{%
  Steffen~Knoblauch \thanks{Contributed equally}\\
  Heidelberg University\\
  \texttt{steffen.knoblauch@heigit.org} \\
  \And
  Levi~Szamek $^{*}$\\
  Heidelberg University\\
  \texttt{levi.szamek@heigit.org} \\
   \And
   Iddy Chazua \\
   OpenMap Development Tanzania \\
   Address \\
  \texttt{iddy.chazua@omdtz.or.tz} \\
   \And
   Benedcto Adamu \\
   OpenMap Development Tanzania \\
   Address \\
  \texttt{benedcto.adamu@omdtz.or.tz} \\
   \And
   Innocent Maholi \\
   OpenMap Development Tanzania \\
   Address \\
  \texttt{innocent.maholi@omdtz.or.tz} \\
   \And
   Alexander Zipf \\
   Heidelberg University \\
   Address \\
  \texttt{zipf@uni-heidelberg.de} \\
}

\begin{document}

\maketitle
\title{AI-based Waste Mapping for Addressing Climate-Exacerbated Flood Risk: Insights from Dar es Salaam}

\begin{abstract}
Urban flooding is a growing climate change-related hazard in rapidly expanding African cities, where inadequate waste management often blocks drainage systems and amplifies flood risks. This study introduces an AI-powered urban waste mapping workflow that leverages openly available aerial and street-view imagery to detect municipal solid waste at high resolution. Applied in Dar es Salaam, Tanzania, our approach reveals spatial waste patterns linked to informal settlements and socio-economic factors. Waste accumulation in waterways was found to be up to three times higher than in adjacent urban areas, highlighting critical hotspots for climate-exacerbated flooding. Unlike traditional manual mapping methods, this scalable AI approach allows city-wide monitoring and prioritization of interventions. Crucially, our collaboration with local partners ensured culturally and contextually relevant data labeling, reflecting real-world reuse practices for solid waste. The results offer actionable insights for urban planning, climate adaptation, and sustainable waste management in flood-prone urban areas.
\end{abstract}

\section{Introduction} \label{chap:introduction}
Climate change is amplifying the frequency and intensity of urban flooding, particularly in rapidly growing cities of the Global South, where informal settlements and inadequate drainage infrastructure heighten exposure to stormwater hazards. A critical yet underacknowledged factor in this dynamic is unmanaged solid waste: when waste obstructs drains, culverts, or riverbeds, hydraulic capacity is reduced, increasing the risk of localized flooding—especially during high-intensity or erratic rainfall events. In Sub-Saharan Africa, less than half of municipal solid waste (MSW) is formally collected, with up to 70\% disposed of through open dumping \cite{Kaza.2018}. These practices impair drainage systems and contribute to broader environmental and public health risks, while undermining urban resilience to climate-related hydrometeorological extremes \cite{Moritz.2023, Carbery.2018}. As global MSW generation is expected to grow from 2.01 to 3.40 billion tonnes by 2050 \cite{Kaza.2018}, scalable monitoring tools are urgently needed. Deep learning applied to high-resolution imagery offers a promising solution for identifying waste hotspots with the granularity required for targeted interventions. While prior work using satellite imagery \cite{Sun.2023, Devesa.2021} has made progress, it lacks the spatial resolution to capture small, dispersed waste piles. UAV and street-level imagery provide richer detail \cite{Wang.2024, Knoblauch.2024, Cinnamon.2024}, but most studies to date focus on technical improvements rather than operational deployment \cite{Abdu.2022}. This paper introduces the first end-to-end AI workflow that integrates UAV and 360° street-view (SV) imagery for large-scale waste detection and flood-related clogging risk assessment. Validated in Dar es Salaam, our approach supports local climate adaptation by identifying high-risk areas where accumulated waste threatens drainage functionality.

 
\section{Materials and methods}\label{materials_and_methods}

Dar es Salaam, Tanzania’s largest city, exemplifies the nexus of rapid urbanization, climate-exacerbated flood risk, and inadequate waste infrastructure. Home to over 5.38 million people in 2022 and projected to surpass 10 million by 2030, the city generates an estimated 4,600 tons of solid waste daily, of which less than 40\% is formally collected \cite{TheUnitedRepublicofTanzaniaURTMinistryofFinanceTanzaniaNationalBureauofStatistics.2024, WorldBank.2024}. Over 70\% of residents live in informal settlements, often in flood-prone areas with limited access to drainage or waste services \cite{Nyyssoa.2021}. Uncollected waste frequently accumulates in waterways and drainage systems, contributing to near-annual floods worsened by increasingly intense rainfall and blocked infrastructure \cite{Brownell.2020, Sakijege.2019}. The Msimbazi River basin, among other low-lying areas, is particularly affected. This setting presents a critical testbed for scalable AI-driven waste detection and flood-risk mitigation tools.

\begin{figure}[H]
\centering
\includegraphics[width=1\textwidth]{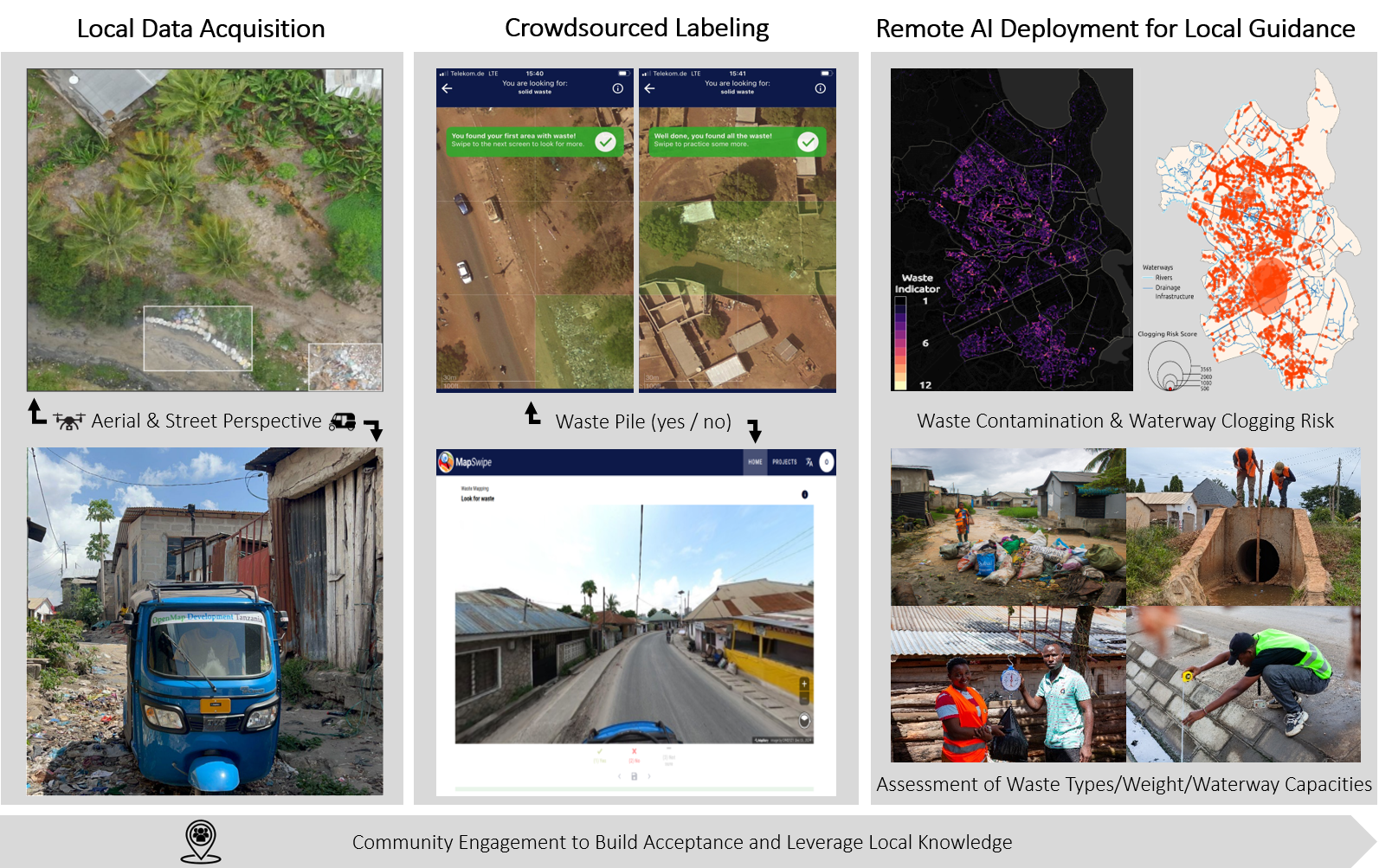}
\caption{Overview of the workflow: (i) in-field data acquisition by local partners, including city-wide UAV flights and street-level image capture; (ii) crowdsourced annotation to embed local knowledge on MSW; and (iii) AI-based detection of waste to generate city-scale clogging risk maps, supporting climate adaptation and urban waste management strategies.}
\label{fig:workflow}
\end{figure}

To obtain a comprehensive, multi-perspective view of MSW, we conducted a two-week data collection campaign beginning October 25, 2023, using both aerial and ground-level imagery. High-resolution UAV data were acquired using a DJI Mavic 2 Pro drone flown at 150 meters altitude, equipped with RTK GPS and ground control points. This setup produced orthomosaics covering 19.2 km² at an average ground sampling distance of ~9 cm. The UAV orthomosaics were tiled into 640×640 pixel patches for downstream model training. For ground-level imagery, a GoPro Max 360° camera was mounted on a tricycle and supplemented with handheld capture in inaccessible areas. Images were recorded every 5 seconds along 1,500 km of streets, yielding over 494,000 panoramic images \cite{Chazua.2025}. The UAV imagery was uploaded to OpenAerialMap  \cite{OpenImageryNetwork.2025}, and the street-view (SV) data to Panoramax \cite{PanoramaxDevelopers.2025}, ensuring open access for downstream training and evaluation. To prepare the 4096×2048-pixel equirectangular SV images for deep learning, each was split into 12 overlapping planar patches, preserving spatial continuity and improving detectability of small objects near image poles. This preprocessing step acted as both a distortion correction strategy and a form of implicit data augmentation. Waste piles were defined as visible, multi-item accumulations of improperly contained solid waste. Isolated litter was excluded. Manual annotation was performed via MapSwipe \cite{MapSwipeDevelopers.2023}, enabling local contributors to embed contextual knowledge. This was essential to distinguish functional materials (e.g., bins for rainwater harvesting, stacked tires used as planters) from actual waste. Such items, being maintained and reused, were assumed to pose negligible clogging or flood risks. In total, 1,691 UAV image patches and 15,415 SV patches were annotated. SV images were sampled every 10 meters from one-third of surveyed paths across city wards, ensuring geographic and contextual diversity.

We fine-tuned a YOLOv6-S model using 1,691 annotated UAV patches. Data augmentation included random cropping, flipping, and brightness/contrast adjustment, producing 4,447 additional samples. We added 307 background-only images to improve generalization, resulting in 6,445 total patches. A 70/15/15 train/test/validation split ensured robust evaluation. Training was performed on a NVIDIA L4 GPU (24 GB VRAM), using a batch size of 16 over 605 epochs with early stopping. An Intersection-over-Union (IoU) threshold of 0.65 and confidence threshold of 0.449 were used during inference. The model was applied to 66,481 UAV patches from the full study area, with detections georeferenced via centroid extraction for downstream mapping. SV detection used a fine-tuned YOLOv11m-cls model \cite{Khanam.2024}. Sky pixels were masked, reducing image height by ~50\%, enhancing focus on ground-level features. We applied environment-specific augmentation (e.g.,  RandomSunFlare, RandomRain, RandomShadow) and geometric perturbations (horizontal flips, slight rotations). Images were resized with aspect ratio preserved and padded as needed. A 70/15/15 data split was performed using spatial KMeans clustering to avoid geographic leakage. Negative examples were downsampled by 20\% in the validation set to emphasize positive class learning.  Training involved reduced learning rate, early stopping, and increased dropout to prevent overfitting. The trained model was deployed on 1,514,208 street-level image patches for large-scale prediction. Model performance was evaluated using the F1-score. 


To produce a unified, city-scale waste accumulation index, we spatially aggregated model predictions from UAV and SV imagery using a shared hexagonal grid with 308 m² cells. For UAV-based detections, the total predicted waste area within each hexagon was normalized by the proportion of UAV coverage in that cell, accounting for partial imagery. For SV-based detections, each SV location contributed a score from 0 to 12, corresponding to the number of positive classifications across its 12 split image patches. These scores were then normalized by the number of SV image points intersecting each hexagon to produce a comparable waste intensity measure. To evaluate spatial correspondence between UAV- and SV-derived waste detections, we applied Local Moran’s I \cite{Moran.1950}, a local spatial autocorrelation measure. This analysis revealed statistically significant clusters of agreement—high-high (HH) and low-low (LL) areas—indicating locations where both modalities consistently detected high or low waste presence. It also identified discordant clusters—high-low (HL) and low-high (LH)—highlighting regions where waste was prominent in one modality but not the other. These patterns, visualized via bivariate spatial maps, provide insight into how waste visibility varies across aerial and ground-level perspectives.

To assess the interplay between solid waste accumulation and flood vulnerability, we overlaid predicted waste locations with hydrological flow models generated from a 30-meter Shuttle Radar Topography Mission (SRTM) Digital Elevation Model (DEM). Using the D8 flow direction algorithm, we delineated surface runoff pathways and computed the Strahler stream order to approximate each waterway's relative capacity to accumulate and convey water. Clogging risk was estimated as the product of the normalized quantity of waste within a fixed buffer around each drainage segment and its corresponding Strahler index, divided by the estimated drainage capacity. Drainage capacity was derived from OpenStreetMap (OSM) data and field surveys conducted in collaboration with a local mapping organization. This approach enabled spatially explicit identification of potential blockage hotspots, incorporating both waste presence and hydrological relevance.

\section{Results and discussion}
\begin{figure}[H]
\centering
\includegraphics[width=1\textwidth]{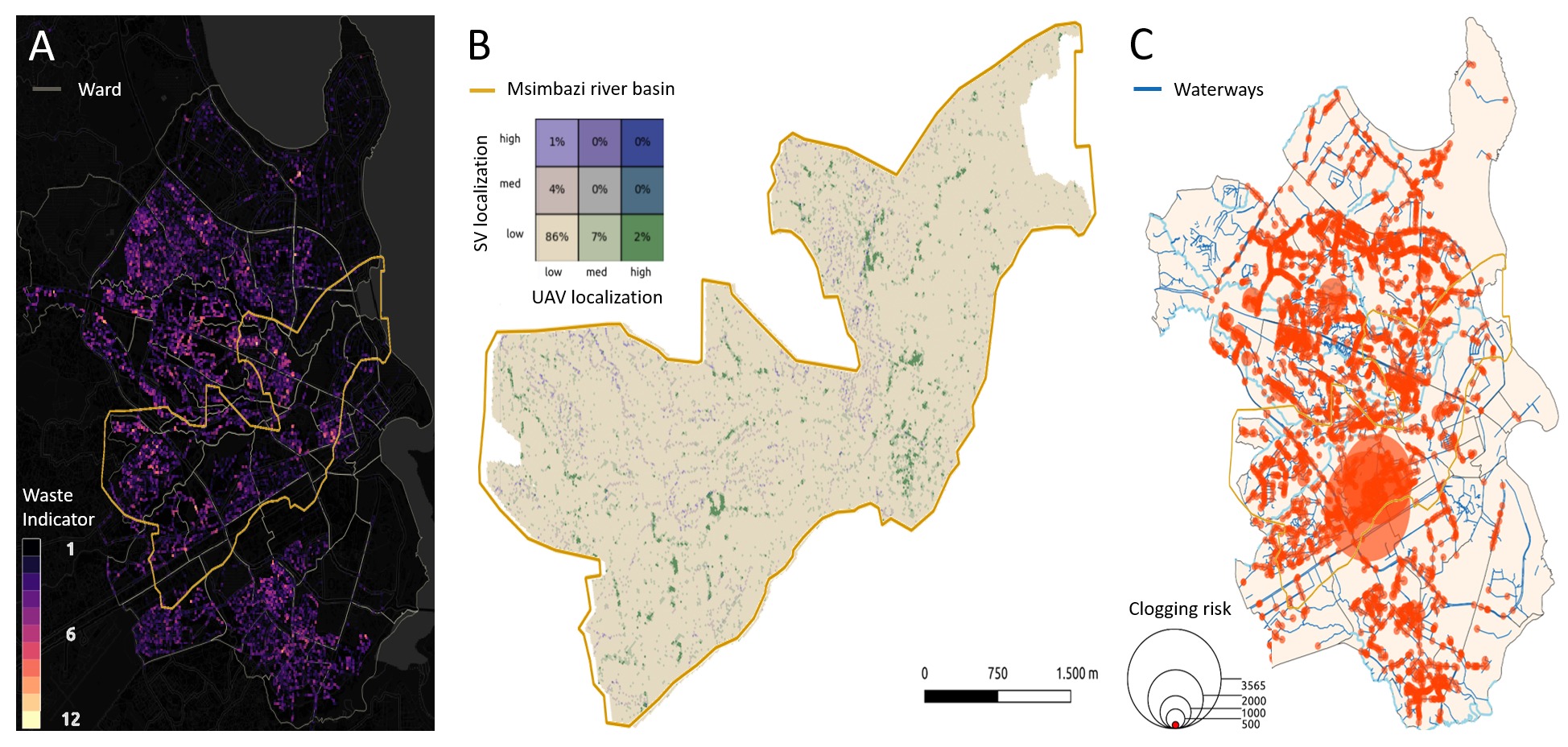}
\caption{
Spatial distribution of detected solid waste and clogging risk implications. 
\textbf{(A)} UAV- and SV-derived waste detections are concentrated in central districts, particularly in informal settlements with high population density and limited waste management infrastructure, while coastal areas exhibit fewer detections, corresponding to higher-income neighborhoods. 
\textbf{(B)} Cross-modal Moran’s I cluster analysis of the Msimbazi River basin reveals minimal spatial overlap between UAV and SV detections—only 0.5\% of hexagons contain both—underscoring the complementary nature of the two perspectives.
\textbf{(C)} Estimated clogging risk at drainage nodes, with circle size proportional to the risk score, reveals high-risk zones concentrated around the Msimbazi river basin and its tributaries.}
\label{fig:waste_map}
\end{figure}

Both models achieved strong performance, with F1 scores of 0.97 (SV) and 0.92 (UAV). UAV imagery had more false positives, often due to reflective water, laundry, mosaicking artifacts, and debris. Waste was concentrated three times more near riverbeds, highlighting flood risks from clogged drainage. UAV and SV data provided complementary insights with limited spatial overlap (Fig. \ref{fig:waste_map}). UAV detected backyard piles well, while SV excelled at smaller or sheltered piles. However, SV coverage is limited to accessible streets and cannot capture areas without roads. On the other hand, SV images also enable estimation of waste pile height and volume, which could enhance future studies by differentiating waste amounts and types. 

\section{Conclusion}
This study highlights the potential of AI-driven waste mapping to support climate adaptation by revealing flood-exacerbating waste accumulation patterns. The combination of UAV and SV imagery was essential to capture spatial variability across Dar es Salaam, uncovering waste hotspots that would be missed using a single modality. Collaboration with local partners was crucial for (i) reducing false positives - such as drying clothes or tires repurposed as planters, (ii) identifying dumping practices undetectable by our methods, like waste concealed beneath riverine vegetation, and (iii) mapping drainage and canal capacities critical for clogging risk analysis. These insights highlight the importance of local knowledge for accurate clogging risk assessment related to solid waste. Through open data sharing, we aim to facilitate the transfer and scaling of this low-cost approach to other flood-prone urban areas. Overall, our results can inform targeted cleanup, infrastructure planning, and climate adaptation strategies aligned with SDGs 3, 6, 11, and 13.

\bibliographystyle{unsrt}
\bibliography{solid_waste_Tanzania}

\end{document}